%% file: main.tex
\documentclass[10pt,twocolumn,letterpaper]{article}

\usepackage{cvpr}              
\usepackage{makecell}
\usepackage{multirow, booktabs}
\usepackage{threeparttable}
\usepackage[accsupp]{axessibility} 

\input{preamble}

\definecolor{cvprblue}{rgb}{0.21,0.49,0.74}
\usepackage[pagebackref,breaklinks,colorlinks,citecolor=cvprblue]{hyperref}
\usepackage{graphicx}
\def\paperID{6270} 
\def\confName{CVPR}
\def\confYear{2024}

\title{Generalizable Whole Slide Image Classification with Fine-Grained Visual-Semantic Interaction}

\author{
\textbf{~~~~~Hao Li$^{1}$}
~~~~~~~~~~\textbf{Ying Chen$^{1}$}
~~~~~~~~~~\textbf{Yifei Chen$^{2}$}
~~~~~~~~~\textbf{Wenxian Yang$^{3}$} \\
\textbf{Bowen Ding$^{4}$} 
~~~~\textbf{Yuchen Han$^{4}$\footnotemark[1]}
~~~~\textbf{Liansheng Wang$^{1}$\footnotemark[1]}
~~~~\textbf{Rongshan Yu$^{1}$\footnotemark[1]}\\
$^{1}$School of Informatics, Xiamen University 
~~~~~~~$^{2}$Huawei 
~~~~~~~$^{3}$Aginome Scientific, Xiamen, China \\
$^{4}$Department of Pathology,  Shanghai Chest Hospital, Shanghai Jiao Tong University School of Medicine\\
{\tt\small \{lllih, cying2023\}@stu.xmu.edu.cn, chenyifei14@huawei.com, wx@aginome.com,} \\
{\tt\small dingbowenmail@126.com, ychan@cmu.edu.cn, \{lswang, rsyu\}@xmu.edu.cn}
}

\begin{document}
\maketitle
\begin{abstract}

Whole Slide Image (WSI) classification is often formulated as a Multiple Instance Learning (MIL) problem. Recently, Vision-Language Models (VLMs) have demonstrated remarkable performance in WSI classification. 
However, existing methods leverage coarse-grained pathogenetic descriptions for visual representation supervision, which are insufficient to capture the complex visual appearance of pathogenetic images, hindering the generalizability of models on diverse downstream tasks. Additionally, processing high-resolution WSIs can be computationally expensive. 
In this paper, we propose a novel ``Fine-grained Visual-Semantic Interaction" (FiVE) framework for WSI classification. It is designed to enhance the model's generalizability by leveraging the interaction between localized visual patterns and fine-grained pathological semantics.
Specifically, with meticulously designed queries, we start by utilizing a large language model to extract fine-grained pathological descriptions from various non-standardized raw reports. 
The output descriptions are then reconstructed into fine-grained labels used for training. 
By introducing a Task-specific Fine-grained Semantics (TFS) module, we enable prompts to capture crucial visual information in WSIs, which enhances representation learning and augments generalization capabilities significantly.
Furthermore, given that pathological visual patterns are redundantly distributed across tissue slices, we sample a subset of visual instances during training.
Our method demonstrates robust generalizability and strong transferability, dominantly outperforming the counterparts on the TCGA Lung Cancer dataset with at least 9.19\% higher accuracy in few-shot experiments.
The code is available at: \href{https://github.com/ls1rius/WSI\_FiVE}{https://github.com/ls1rius/WSI\_FiVE}.
\end{abstract}  


\renewcommand{\thefootnote}{\fnsymbol{footnote}} 
\vspace{-0.8em}
\footnotetext[1]{Co-corresponding authors. This work was supported by National Natural Science Foundation of China (Grant No. 62371409).}


\section{Introduction}
\label{sec:intro}

\begin{figure}[th]
  \centering
   \includegraphics[width=\linewidth]{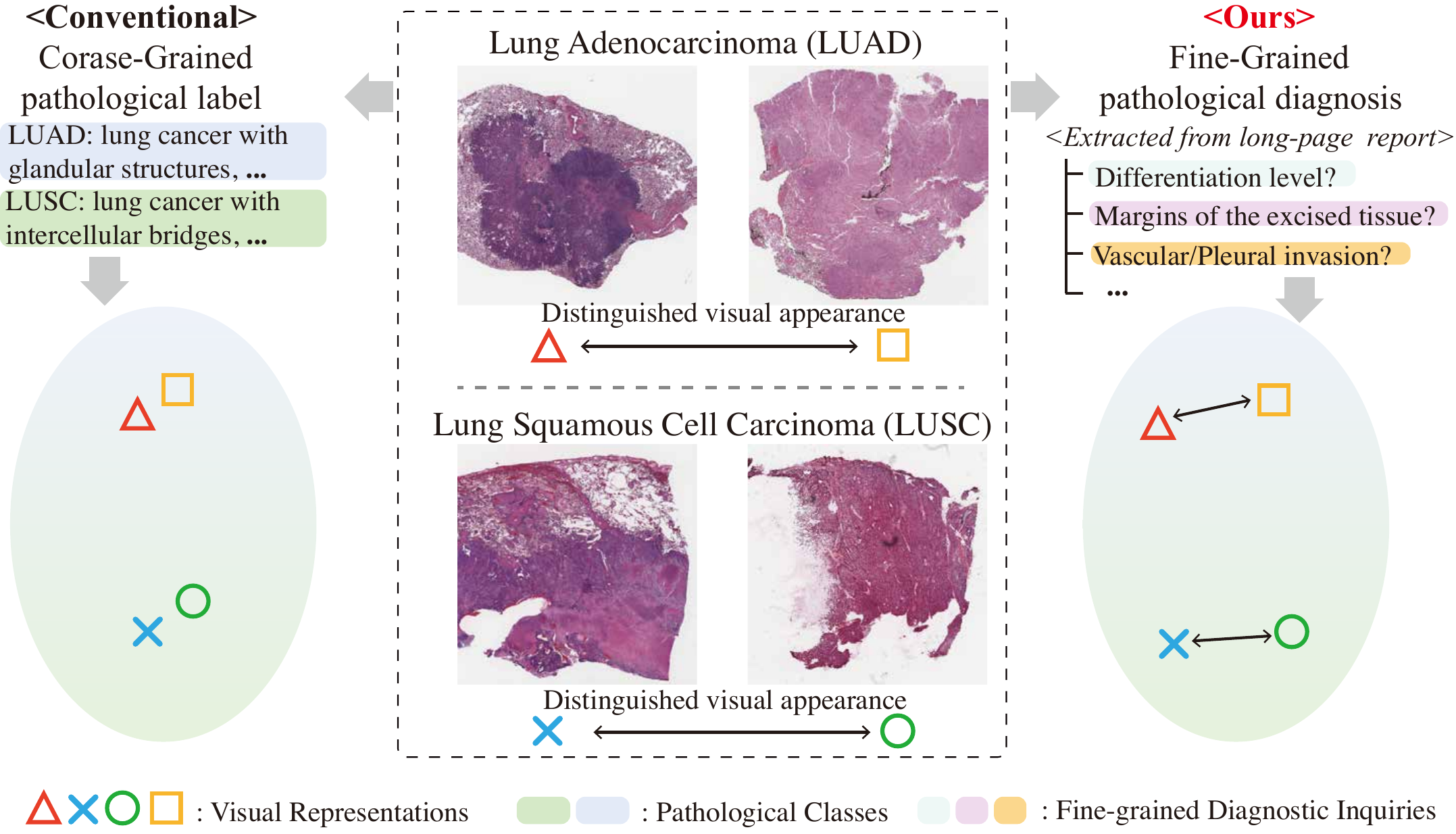}
   \caption{Challenges in WSI-text contrastive learning. Most conventional VLM approaches categorize whole slide images using category-level text descriptions, overlooking intra-class differences, leading to a decline in performance and limitations in generalization capabilities. Instead, we extract fine-grained descriptions from pathology reports as slide-level labels to develop our model, exhibiting detailed variations in each WSI.}
   \label{fig_main_problem}
\end{figure}

Histological Whole Slide Image (WSI) classification plays a crucial role in computational pathology by automating disease diagnosis and subtyping. For high-resolution WSI analysis, Multiple Instance Learning (MIL) has become the dominant method. These methods treat each WSI as a ``Bag'' sample, and aggregate numerous patches within it as instances for thorough decision-making. Nevertheless, most existing methods~\cite{lin2023interventional,li2023task, xiang2022exploring, li2021dual} focused on processing image data to conduct WSI classification, potentially not emphasizing critical pathological insights, particularly the expert textual annotations that accompany these slides. 

Recently, Vision-Language Models (VLMs)~\cite{radford2021learning, jia2021scaling, yao2021filip} underscored the significance of integrating multimodal information for developing robust encoders. Zhang et al.~\cite{zhang2023text} exploited disease-grade text labels and extracted text insights using pre-trained language models. Qu et al.~\cite{qu2023rise} utilized GPT-4 in a question-and-answer mode to obtain language prior knowledge at both instance and bag levels for VLM training. However, the challenge lies in the uniqueness and variability of content in each WSI. Existing methods developed their models with coarse-grained descriptions (\emph{i.e.}, simplistic Category-Level text labels~\cite{zhang2023text} or descriptive Category-Level text labels constructed by GPT-4~\cite{qu2023rise}), as shown in ~\cref{fig_main_problem}. They may have omitted crucial fine-grained pathological details, including differentiation level, vascular invasion, etc., which results in reduced model performance and limited generalization.

WSIs accompanied by their corresponding reports (\emph{i.e.}, WSI-report pairs) offer detailed descriptions and fine-grained information vital for WSI analysis. Furthermore, a substantial collection of these pairs is accessible in public databases, such as The Cancer Genome Atlas (TCGA)~\cite{hutter2018cancer}. However, their full potential has not been adequately harnessed yet. The challenge in developing a Visual Language Model (VLM) using WSI-report pairs mainly lies in the diverse formats and standards of the raw reports from different hospitals, which increases the complexity of data preprocessing and standardization processes. Additionally, pathology reports often contain extraneous information, including report metadata, tissue processing descriptions, and repetitive elements, which can introduce noise to the textual data. \textbf{\emph{How to extract useful information from raw pathology reports to construct WSI-report pairs}} is a key issue. Moreover, recent studies have demonstrated the efficacy of prompt engineering in enhancing VLMs. In contrast to natural images, WSI data encompasses extensive professional pathological information and intricate details. \textbf{\emph{How to craft prompts to make full use of this semantic information to guide fine-grained feature learning}} is a challenging task. Besides, the high computational costs to train models with high-resolution WSIs also limits the promotion of the model, resulting in a certain resource threshold for WSI analysis. 
 
To address these issues, we propose a novel whole slide image classification method with \textbf{Fi}ne-grained \textbf{V}isual-s\textbf{E}mantic interaction termed as FiVE, which shows robust generalizability and efficiency in computation. Firstly, we obtain WSIs with non-standardized raw pathology reports from a public database. Collaborating with professional pathologists, we craft a set of specialized prompts to standardize reports. Following this, we employ the large language model GPT-4 to automatically clean and standardize the raw report data. In addition, we propose the \textbf{T}ask-specific \textbf{F}ine-grained \textbf{S}emantic (TFS) Module, which utilizes manual-designed prompts to direct visual attention to specific pathological areas while constructing Fine-Grained Guidance to enhance the semantic relevance of model features. Considering the diffuse distribution of pathological diseases within tissue sections and the presence of numerous non-diagnostic regions in WSIs, we also incorporate a patch sampling strategy during the training phase to enhance training efficiency and reduce computational costs. The contributions of this paper are summarized as follows:

\begin{itemize}
\item[$\bullet$] We pioneer the utilization of the available WSI diagnostic reports with fine-grained guidance. The obtained fine-grained description labels lead to improved supervision by discriminating the visual appearances more precisely.

\item[$\bullet$] We introduce a novel Task-specific Fine-grained Semantics (TFS) Module to offer fine-grained guidance, significantly enhancing the model's generalization capabilities.

\item[$\bullet$] We implement a patch sampling strategy on visual instances during training to enhance computational efficiency without significantly compromising accuracy, thereby optimizing the model's training process.

\end{itemize}

\section{Related Work}
\subsection{Whole Slide Image Analysis}
Contemporary methodologies for WSI analysis predominantly employ MIL methods where each WSI is treated as a ``Bag'' and its extracted patches as instances within this bag. MIL methods consists of instance-based methods~\cite{campanella2019clinical, lerousseau2020weakly, xu2019camel} and embedding representation-based methods~\cite{li2021dual, lu2021data, shao2021transmil, zhang2022dtfd, wang2018revisiting}. However, the majority of existing methods~\cite{shao2023lnpl, qu2023boosting, lin2023interventional, chan2023histopathology} almost exclusively rely on image data, neglecting vital pathological details, notably the specialist text annotations that accompany the images. Recent works~\cite{zhang2023text, qu2023rise} have taken note of this issue and started to utilize text information to improve pathological image classification. They used bag-level text labels or the descriptive labels generated by GPT. However, given the unique and varied descriptions of each WSI, their methods fall short of fully leveraging the detailed textual information present in the slides.

\begin{figure*}[h]
  \centering
   \includegraphics[width=\linewidth]{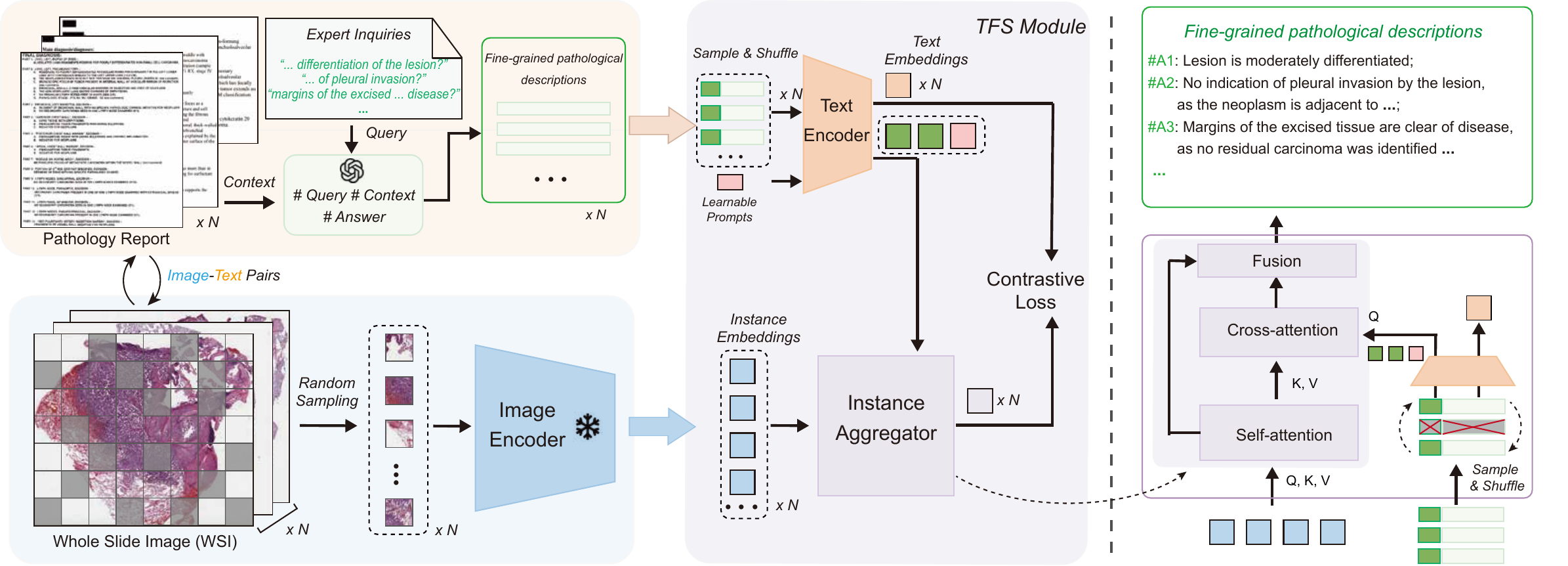}
   \caption{\textbf{Left: The structure of the FiVE framework.} The model consists of a frozen image encoder, a text encoder, and the TFS module. Whole slide images are divided into instances for embedding extraction by the image encoder. Raw pathology reports are standardized by GPT-4 into fine-grained descriptions. The fine-grained descriptions and manual prompts are sampled, shuffled, and reconstructed in pairs. These prompts aggregate instances into bag-level features, subsequently aligned with the descriptions utilizing contrastive loss. \textbf{Top Right: Fine-grained pathological descriptions.} The fine-grained pathological descriptions are generated from multiple answers based on specific queries. These descriptions undergo a process of random sampling, shuffling, and reconstruction to form a unified sentence. \textbf{Bottom Right: The Instance Aggregator module.} The instance aggregator consists of a self-attention module and a cross-attention module, fusing image instance embeddings and prompt embeddings to create bag-level features.}
   \label{fig_main}
\end{figure*}

\subsection{Vision-Language Models}
Recent researches have made efforts to develop Vision-Language Models (VLMs). CLIP~\cite{radford2021learning} gathered 400 million image-text pairs and initiated training with synchronized vision and text encoders from the onset. LiT~\cite{zhai2022lit} developed a text encoder compatible with a pre-trained vision encoder. FLIP~\cite{yao2021filip} integrated region-word alignment in contrastive learning to enhance detailed vision-language correlation. Coca~\cite{yu2022coca} pre-trained an image-text encoder-decoder foundation model using contrastive and captioning loss. 
For pathological images, some research works adapted VLMs for training with pathological images and text. Lu et al.~\cite{lu2023towards} built a VLM using over 1.17 million histopathology image-caption pairs based on a task-agnostic pre-training approach derived from Coca. Huang et al.~\cite{huang2023visual} curated the OpenPath dataset, consisting of 208,414 pathology images with natural language descriptions from public forums, and fine-tuned a pre-trained CLIP on OpenPath. Lai et al.~\cite{lai2023clipath} also explored the generalization of CLIP in pathology image classification. These methods typically employ instance-level images (small patches from WSIs) and descriptions, requiring significant human and material resources. Zhang et al.~\cite{zhang2023text} injected meaningful medical domain knowledge to advance pathological image embedding and classification. Qu et al.~\cite{qu2023rise} employed GPT-4 to supplement image labels to enrich the information for training. However, the texts employed in their training offer only rudimentary descriptions of the images, primarily categorizing the general type of pathological slides, such as the disease category. This approach significantly constrains the model's capacity to discern fine-grained features, including the degree of differentiation, spread, and other details. Consequently, this limitation substantially restricts the model's generalizability and applicability to more nuanced diagnostic tasks.

\subsection{Prompt Learning in Vision-Language Models}
Drawing inspiration from prompt learning in natural language processing, some studies have proposed adapting Vision-Language models through end-to-end training of prompt tokens. CoOp~\cite{zhou2022learning} enhanced CLIP for few-shot transfer by optimizing a continuous array of prompt vectors within its language branch. CoCoOp~\cite{zhou2022conditional} identified CoOp's suboptimal performance on new classes and tackled the generalization issue by conditioning prompts directly on image instances. Lu et al.~\cite{lu2022prompt} advocated for optimizing diverse sets of prompts by understanding their distribution. Bahng et al.~\cite{bahng2022visual} undertook visual prompt tuning on CLIP, focusing the prompting on the vision branch. MaPLe~\cite{khattak2023maple} investigated the effectiveness of multi-modal prompt learning in order to improve alignment between vision and language representations. Zhang et al.~\cite{zhang2023llama} adopted a set of learnable adaption prompts and prepend them to the word tokens at higher transformer layers, efficiently fine-tuning LLaMA with less cost. Furthermore, in the context of WSI classification, prompts function as valuable adjuncts, enriching contextual information and semantic interpretation. The strategic utilization of prompts substantially improved model performance~\cite{qu2023rise}.

\begin{table*}
  \centering
  \resizebox{\linewidth}{!}{
  \begin{tabular}{c|c}
    \toprule
    Manual-Designed Standards & Fine-Grained Text Description Label Examples \\
    \midrule
    \makecell[l]{1. What is the differentiation of the lesion?} & \multirow{3}{*}{\makecell[l]{\textbf{TCGA-44-6774}: Lesion differentiation is moderately to poorly\\ differentiated; Unknown; No indication of vascular invasion by\\ the lesion; No indication of pleural invasion by the lesion;\\ Unknown; Margins of the excised tissue are clear of disease. }} \\
    \makecell[l]{2. Is there any indication of spread through air spaces \\around the lesion?} &\\
    \makecell[l]{3. Is there any indication of vascular invasion by the lesion?} &\\
    \makecell[l]{4. Is there any indication of pleural invasion by the lesion?} &\multirow{3}{*}{\makecell[l]{\textbf{TCGA-49-4505}: Lesion differentiation is well-differentiated;\\ Unknown; Unknown; Pleural invasion by the lesion is present,\\ as the carcinoma extends through the visceral pleura; The lesion\\ invades adjacent tissues or organs; Margins of the excised tissue\\ are clear of disease.}}\\
    \makecell[l]{5. Is there any evidence of the lesion invading adjacent\\ tissues or organs?} &\\
    \makecell[l]{6. Are the margins of the excised tissue clear of disease?\\} &\\
    &\\
    \bottomrule
  \end{tabular}
  }
  \caption{Manual-Designed Standards and Fine-Grained Text Description Label Examples. The answers on the right correspond to the standards on the left. ``Unknown" is used as a placeholder when relevant information cannot be found.}
  \label{tab_human_prompt}
\end{table*}

\section{Method}
\subsection{Overview}

\cref{fig_main} shows the pipeline of our proposed FiVE method. 
To initiate the process, we collaborate with professional pathologists to establish a set of standards. Following this, we employ GPT-4 to automatically extract and standardize information based on these various standards.
During the training phase, we construct Fine-Grained Guidance by intricately dividing and reconstructing the text description labels and manual prompts in pairs. 
The combination of manual prompts and learnable prompts forms the Diagnosis Prompts, which are utilized to enhance the semantic relevance of the features.
Subsequently, the instance aggregator module fuses instance features with fine-grained prompts, generating bag-level features, subsequently align with the corresponding fine-grained text description labels.
Additionally, to reduce computational costs, we implement the patch sampling strategy, optimizing the model’s training efficiency while minimizing performance loss.

\subsection{Text Standardization via GPT-4}
We utilize fine-grained text description labels extracted from pathology reports to align image bag-level features. Though pathology reports are readily accessible from public databases, their content exhibits significant variability depending on the source. Despite these format differences, pathology diagnoses consistently adhere to specific and well-established diagnostic standards. In our work, we develop fine-grained diagnostic criteria under the guidance of professional pathologists to standardize report data and extract fine-grained insights pertinent to pathological diagnosis, aiming to enhance its generalization capabilities substantially.

The manual-designed standards aim to extract the morphological characteristics under the microscope, such as the degree of differentiation and lesion invasion, and filter out information irrelevant to the diagnosis.
Subsequently, we employ GPT-4 to automatically extract answers from the original diagnosis reports based on prompts composed of these standards. If the information queried is absent in the pathology reports, ``Unknown'' is used as the answer. \cref{tab_human_prompt} shows manual-designed standards and two fine-grained text description label examples. 
Then, we recombine these extracted fine-grained information and integrate it into a complete description of the case image. \textbf{More details about the prompts used for Text Standardization are provided in Supplementary Material.}

\subsection{Task-specific Fine-grained Semantics Module}

\subsubsection{Fine-grained Guidance Construction}
Due to the Text Standardization process, our data has achieved standardization. 
Utilizing these fine-grained text description directly as training labels can yield performance improvements. Additionally, leveraging them to generate more diverse and semantically enriched fine-grained guidance can further boost the model's performance.

During the training process we utilize these manual-designed standards as our manual-designed prompts. We divide the original fine-grained text descriptions into several parts according to manual-designed prompts, followed by random sampling and eliminating ``Unknown'' tags from the initial labels. Given that the staged diagnostic reports in pathological descriptions are sequence-independent, we shuffle these preliminary labels and reconstruct them into a full-sentence description. During the training, we train reconstructed text description labels and reconstructed manual prompts in pairs. For example, consider description A: \textbf{\emph{``Lesion differentiation is moderately differentiated; Margins of the excised tissue are clear of disease.''}} and description B: \textbf{\emph{``Lesion differentiation is moderately differentiated; Margins of the excised tissue are not clear of disease, as the tumor is within the bronchial margin and parenchymal margin.''}}. When only the first part of each description is sampled, they would be grouped into the same category. However, when sampling the entire sections, they are considered as distinct descriptions. Changes in granularity provide diverse perspectives on the visual image, aligning visual image with text descriptions of varying granularities.

This strategy offers three key benefits: 1) Effectively alleviating the parent-child relationship in pathology categories. 2) Providing additional hierarchical semantic perspectives to enhance the text encoder's semantic comprehension ability. 3) Mitigating discrepancies in category annotation due to incomplete diagnostic information.

\subsubsection{Diagnosis Prompts}
We introduce Diagnosis Prompts to guide the aggregation of instance features into bag-level features. We compute the similarity between the instance features and the given manual prompts, utilizing the similarity scores as weights $W$ for feature aggregation to improve the task-specific relevance of the features. Here we utilize the identical manual prompts as those used to standardize the raw data, as shown in the left of the~\cref{tab_human_prompt}.

In addition, manual-designed prompts may have some flaws, potentially failing to comprehensively capture the specific morphological characteristics of the lesion, and the model struggles to generalize towards unseen classes due to the late fusion through the transformer layers.
Besides, fine-tuning the model may not always be feasible as it requires training a large number of parameters. Particularly in the case of low-data regimes, where the availability of training data like whole slide images is extremely limited. 
LLaMA-Adapter~\cite{zhang2023llama} and LLaMA-Adapter-v2~\cite{gao2023llama} explore the way to efficient fine-tuning of Language Models and Vision-Language Models respectively. These approaches introduced the Adaptation Prompt to gradually acquire instructional knowledge. They adopted zero-initialized attention with gating mechanisms to ensure stable training in the early stages.
Inspired by these methods, we introduce learnable continuous diagnosis prompts to enrich the context information and enhance the model's transferability.

Specifically, we get the manual text prompt tokens $Q_h=[q_{h1}, q_{h2}, \cdots, q_{hn}]$, here $n$ represents the number of the manual prompts. We concatenate the learnable continuous prompt tokens $Q_{l}=[q_{l1}, q_{l2}, \cdots , q_{lm}]$ on it, here $m$ represents the number of the learnable prompt tokens. Finally we get the diagnosis prompts $Q=[q_{h1}, q_{h2}, \cdots , q_{hn}, q_{l1}, q_{l2}, \cdots , q_{lm}]$. In the training phase, part of manual prompt tokens $Q_h$ will be sampled randomly paired with text description labels, while the whole learnable prompt tokens $Q_l$ will be consistently retained. 

Different with the traditional context learning prompts method, our approach pays attention to the acquisition of prior knowledge, similar to the methodology employed in Detection Transformer (DETR)~\cite{carion2020end}. We aim to acquire a set of appropriate query values to improve performance in subsequent feature screening processes. Additionally, it can also enable the model to quickly transfer to other tasks by fine-tuning this set of queries.

\subsubsection{Instance Aggregator Module}
The Instance Aggregator (IA) module is used to aggregate the fine-grained diagnosis prompts and instance features. As shown in the right of the ~\cref{fig_main}, IA consists of a self-attention module and a cross-attention module. 

We employ self-attention to enable feature interaction among instance features $I_{i} = [e_{i1}, e_{i2}, \cdots, e_{ij}]$, resulting in the feature $s_{i}$. Subsequently, utilizing the diagnosis prompts $Q$ to aggregate the instance features and acquire the feature $z_{i}$. Then we concatenate $s_{i}$ and $z_{i}$, utilizing the learnable parameter $W$ to fuse these features, yielding the bag-level feature $v_{i}$. The formulas are shown as follows:

\vspace{-1.0em}
\begin{align}
s_{i} &= SelfAttention(I_{i}, I_{i}) + I_{i} \\
z_{i} &= CrossAttention(Q, s_{i}) \\
v_{i} &= concat(mean(s_{i}), mean(z_{i})) \cdot W
\end{align}
 
Ultimately, we acquire the image bag-level features guided by the fine-grained diagnosis prompts, which are then employed to align the fine-grained text features.

\subsection{Patch Sample Strategy}
Each Whole Slide Image (WSI) is partitioned into a variable number of instances, ranging from approximately 50 to 45,000. Handling such a wide range of instances markedly increases computational complexity and substantially extends the training duration.
FLIP~\cite{yao2021filip} reduced computation and reached higher accuracy than CLIP~\cite{radford2021learning} counterpart by randomly removing a large portion of image patches during training~\cite{li2023scaling}. In the case of whole slide images, pathological visual patterns are often redundantly distributed across a tissue slice. Therefore, it is feasible to sample only a subset of visual instances during training. 

For the instances in each bag (\emph{i.e.}, slide), we select a sample amounting to $S_m$ percent of the total number $p$ of each group of instances (patches). The required number of instances $S_n$ is described by the following formula:
\begin{align}
S_n = min(p * S_m, S_{maxn})
\end{align}

Here $S_{maxn}$ denotes the maximum number of sample instances. To achieve this, we evenly divide each group of instances into $S_n$ chunks and randomly select one instance from each chunk. Since the different whole slide images could sample different $S_n$, we pad the rest of the space with the same padding.

\subsection{Encoder and Loss Function}
We divide each WSI into instances $x_{k}$ and encode these instances into embeddings $e_{k} \in R^{D}$ using pre-trained vision encoder $E_{img}$, composed with ResNet structure following~\cite{li2021dual}, here $D$ represents the dimension of the embeddings.
Then we send the instance embeddings into the TFS Module to aggregate the instance features and prompts, and obtain the bag-level embeddings $v_{i} \in R^{D}$. The formulas are shown as follows:
\begin{align}
e_{k} &= E_{img}(x_{k}) \\
I_{i} &= [e_{i1}, e_{i2}, \cdots, e_{ik}] \\
v_{i} &= IA(I_i, Q)
\end{align}

Besides, we generate pathologically meaningful text embeddings for each WSI, represented as $t_{i} \in R^{D}$, by leveraging the fine-tuned text encoder BioClinicalBERT~\cite{wolf2019huggingface}.
\begin{align}
t_{i}^{c} = E_{txt}(x_{txt}^{c})
\end{align}
where $E_{txt}$ denotes the text encoder, and $x_{txt}^{c} (c~\in[1, C])$ where $C$ denotes the number of categories. Here we use the same embedding dimension $D$ as the vision encoder, suitable for contrastive learning. 
For text encoder $E_{txt}$, we adopt the Low-Rank Adaptation (LoRA)~\cite{hu2021lora} approach for efficient fine-tuning.

Subsequently, the bag-level embeddings $v_{i}$ are aligned with the text embeddings $t_{i}^{c}$ to complete the training process. In this case, prediction $\hat{y}$ is obtained by applying softmax on scaled cosine similarities between the image embeddings and text embeddings:
\begin{align}
	p(\hat{y}=c|I) = \frac{exp(sim(t_{i}^{c}, v_{i})/\tau)}{\sum_{c^{'}=1}^{C}exp(sim(t_{i}^{c^{'}}, v_{i})/\tau)}
\end{align}
where $sim(\cdot, \cdot)$ refers to cosine similarity and $\tau$ is the temperature parameter.

The fine-grained training loss is computed as the cross-entropy between the logits and soft targets as:
\begin{align}
    L^{v \to t} = -\frac{1}{N}\sum_{i=1}^{N}\sum_{i=j}^{N}y_{ij}log(p_{ij})
\end{align}
here $N$ corresponds to the batch size. 

Likewise, we can compute $L^{t \to v}$ and serve $L$ as the final training objective.
\begin{align}
    L = \frac{L^{v \to t} + L^{t \to v}}{2}
\end{align}

\section{Experiments and Results}

\renewcommand{\thefootnote}{\arabic{footnote}}

\subsection{Datasets}
We evaluated our method on public histopathology WSI datasets: The Cancer Genome Atlas Lung (TCGA Lung) Cancer\footnote{\url{http://www.cancer.gov/tcga}} and Camelyon16~\cite{bejnordi2017diagnostic}. 

\noindent \textbf{TCGA Lung Cancer}. The TCGA Lung Cancer dataset comprises two cancer subtypes: Lung Adenocarcinoma (LUAD) and Lung Squamous Cell Carcinoma (LUSC). It includes diagnostic slides with 541 slides from 478 LUAD cases and 512 slides from 478 LUSC cases. For WSI preprocessing, following~\cite{li2021dual}, we cropped each WSI into 256 × 256 non-overlapping patches and removed the background region. The dataset encompasses approximately 5.2 million patches at 20× magnification, averaging about 5,000 patches per WSI. Following~\cite{tang2023multiple}, we randomly split the dataset into training, validation, and testing sets with a ratio of 65:10:25 on the patient level and adopted 4-fold cross-validation. We collected corresponding pathology reports \footnote{\url{https://github.com/tatonetti-lab/tcga-path-reports}} and cleaned them by GPT-4 to produce fine-grained text description labels. To ensure professionalism and accuracy, we invited professional pathologists to check and correct the textual labels. 
Additionally, to evaluate the generalizability of the model and perform zero-shot classification, we utilized a dataset of histological subtype labels for TCGA-LUAD from the cBioPortal database\footnote{\url{https://www.cbioportal.org/}},  \textbf{More details about subtype labels are provided in Supplementary Material.}


\noindent \textbf{Camelyon16}. The Camelyon16 dataset~\cite{bejnordi2017diagnostic} consists of 399 Hematoxylin and Eosin (H\&E) stained slide images, utilized for metastasis detection in breast cancer. We preprocessed each WSI by segmenting it into 256 × 256 non-overlapping patches, excluding background regions. In total, this process yielded approximately 2.8 million patches at a 20× magnification level, with about 7,200 patches per bag. We adopted 3-times 3-fold cross-validation.

\subsection{Implementation Details}
In experiments, we employed ResNet following~\cite{li2021dual} as image encoder to extract image features, while pre-trained BioClinicalBERT from~\cite{wolf2019huggingface} as text encoder to generate text features. LoRA~\cite{hu2021lora} was adopted for fine-tuning the text encoder, with an alpha value of 32 and a rank value of 8. We divided the whole slide image into patches with $256 \times 256$, then applied a random crop with size $224 \times 224$. We adopted AdamW~\cite{loshchilov2017decoupled} with beta (0.9, 0.98), eps 1e-8, learning rate 3e-6, warmup ratio 0.1, weight decay 1e-4 as our optimizer. Additionally, we used the batch size of 1 with an accumulation step of 8 and trained for 150 epochs. We utilized mixed-precision training on 4 NVIDIA-A800 GPUs. 


\subsection{Zero-Shot Histological Subtype Classification}
Prior researches~\cite{qu2023rise, tang2023multiple, zhang2023text} have predominantly concentrated on classifying primary cancer categories. Our approach extends beyond this by attempting to classify detailed histological subtypes. 
It is crucial to emphasize that this task poses a significant challenge, often proving difficult for even skilled pathologists to make direct judgments. 

Since only the LUAD's histological subtype dataset was provided on the online database, we conducted zero-shot subtype classification evaluation on LUAD subtype datasets, with the model being pre-trained on TCGA-LUAD or TCGA-LUSC. 
Throughout the training phase, subtype label information was deliberately excluded, ensuring that all experiments are conducted solely with fine-grained labels. This aims to evaluate the model's ability to identify novel diagnostic categories without specific training on subtypes. We extended the morphological appearance text description labels of the target data using GPT-4. After the text encoder encodes the labels, similarity calculation with image features achieved zero-shot classification.

Since existing zero-shot learning methods cannot be used in WSI subtype classification, we constructed three Linear-Probe method baselines: Mean pooling, Max pooling, and Attention pooling. 
As shown in~\cref{tab_zero_shot}, FiVE attains 65.23\% top-1 accuracy and 95.18\% top-5 accuracy when pre-trained with TCGA-LUAD fine-grained labels. When pre-trained with TCGA-LUSC fine-grained labels, it achieves 62.02\% top-1 accuracy and 94.36\% top-5 accuracy. Moreover, the zero-shot performance of FiVE notably exceeds that of the baseline. This capability in classifying LUAD subtypes is attributed to the focus on fine-grained pathological features during training.

\begin{table}[h]
  \centering
  \begin{tabular}{ccccc}
    \toprule
    \multirow{2}{*}{\begin{tabular}[c]{@{}c@{}}Method\end{tabular} } & \multicolumn{2}{c}{TCGA-LUAD} & \multicolumn{2}{c}{TCGA-LUSC} \\
    \cmidrule(r){2-3} \cmidrule(r){4-5}
    & Top-1 & Top-5 & Top-1 & Top-5 \\
    \midrule
    Mean-pooling & 40.82 & 83.46 & 31.86 & 82.96 \\
    Max-pooling & 45.05 & 88.77 & 36.36 & 86.09 \\
    Attention-pooling & 58.36 & 93.53 & 54.41 & 92.50 \\ \midrule
    FiVE (Ours) & \makecell[r]{\textbf{65.23}\\ \textcolor{blue}{+6.87}} & \makecell[r]{\textbf{95.18}\\ \textcolor{blue}{+1.65}} & \makecell[r]{\textbf{62.02}\\ \textcolor{blue}{+7.61}} & \makecell[r]{\textbf{94.36}\\ \textcolor{blue}{+1.86}} \\
    \bottomrule
  \end{tabular}
  \caption{Zero-Shot performance on histological subtype classification. We pre-trained the model using fine-grained labels of TCGA-LUAD and TCGA-LUSC, then applied zero-shot classification to the histological subtypes of TCGA-LUAD.}
  \label{tab_zero_shot}
\end{table}

\subsection{Few-Shot Classification}
Our model demonstrates adaptability to various tasks even in scenarios with limited data availability. 
Few-shot experiments were conducted to demonstrate its transferability to downstream tasks. 
We initialized the networks with pre-trained weights derived from the model trained on TCGA image-report pairs, and subsequently fine-tuned the model on downstream datasets for few-shot image classification.
We followed ~\cite{qu2023rise} and conducted experiments with 1, 2, 4, 8, 16, and additional 0 shot on the downstream dataset. The results are summarized in ~\cref{tab_finetune}.

\begin{table}[h]
  \centering
  \resizebox{\linewidth}{!}{
  \begin{tabular}{ccccccc}
    \toprule
    Method & 16-shot & 8-shot & 4-shot & 2-shot & 1-shot & 0-shot \\
    \midrule
    \makecell[c]{Mean-pool} & 65.33 & 53.89 & 44.85 & 52.93 & 45.34 & \textbackslash \\ 
    \makecell[c]{Max-pool}  & 48.48 & 49.55 & 44.22 & 48.39 & 49.03 & \textbackslash \\ 
    \makecell[c]{Attn-pool} & 72.50 & 65.79 & 62.47 & 58.36 & 56.23 & \textbackslash \\ 
    \makecell[c]{CoOp~\cite{zhou2022learning}} & 78.35 & 67.99 & 67.60 & 67.54 & 67.81 & \textbackslash \\ 
    TOP~\cite{qu2023rise} & 82.06 & 80.51 & 75.41 & 72.38 & 71.01 & \textbackslash \\ \midrule
    \makecell[c]{FiVE\\ (Ours)} & \makecell[r]{\textbf{91.25}\\ \textcolor{blue}{+9.19}} & \makecell[r]{\textbf{90.80}\\ \textcolor{blue}{+10.29}} & \makecell[r]{\textbf{88.10}\\ \textcolor{blue}{+12.69}}\ & \makecell[r]{\textbf{85.51}\\ \textcolor{blue}{+13.13}} & \makecell[r]{\textbf{83.91}\\ \textcolor{blue}{+12.90}} & \makecell[r]{\textbf{71.26}\\ \textcolor{blue}{}} \\
    \bottomrule
  \end{tabular}
  }
  \caption{Few-shot classification performance on TCGA Lung Cancer. Mean-pool, Max-pool, and Attn-pool correspond to Linear-Probe implementations with Mean-pooling, Max-pooling, and Attention-pooling, respectively.}
  \label{tab_finetune}
\end{table}

\begin{table*}[h]
  \centering
  \begin{tabular}{ccccccc}
    \toprule
    \multirow{2}{*}{\begin{tabular}[c]{@{}c@{}}Method\end{tabular} } & \multicolumn{3}{c}{Camelyon16} & \multicolumn{3}{c}{TCGA Lung Cancer} \\
    \cmidrule(r){2-4} \cmidrule(r){5-7}
    & ACC & AUC & F1-score & ACC & AUC & F1-score \\
    \midrule
    Max-pooling & 78.95$\pm$2.28 & 81.28$\pm$3.74 & 71.06$\pm$2.59 & 81.49$\pm$1.24 & 86.45$\pm$0.71 & 80.56$\pm$1.09 \\
    Mean-pooling & 76.69$\pm$0.20 & 80.07$\pm$0.78 & 70.41$\pm$0.16 & 84.14$\pm$2.97 & 90.13$\pm$2.40 & 83.39$\pm$3.14 \\
    ABMIL~\cite{ilse2018attention} & 90.06$\pm$0.60 & 94.00$\pm$0.83 & 87.40$\pm$1.05 & 88.03$\pm$2.19 & 93.17$\pm$2.05 & 87.41$\pm$2.42 \\
    DSMIL~\cite{li2021dual} & 90.17$\pm$1.02 & 94.57$\pm$0.40 & 87.65$\pm$1.18 & 88.32$\pm$2.70 & 93.71$\pm$1.82 & 87.90$\pm$2.50 \\
    CLAM-SB~\cite{lu2021data} & 90.32$\pm$0.12 & 94.65$\pm$0.30 & 87.89$\pm$0.59 & 87.74$\pm$2.22 & 93.67$\pm$1.64 & 87.36$\pm$2.24 \\
    CLAM-MB~\cite{lu2021data} & 90.14$\pm$0.85 & 94.70$\pm$0.76 & 88.10$\pm$0.63 & 88.73$\pm$1.62 & 93.69$\pm$0.54 & 88.28$\pm$1.58 \\
    TransMIL~\cite{shao2021transmil} & 89.22$\pm$2.32 & 93.51$\pm$2.13 & 85.10$\pm$4.33 & 87.08$\pm$1.97 & 92.51$\pm$1.76 & 86.40$\pm$2.08 \\
    DTFD-MIL~\cite{zhang2022dtfd} & 90.22$\pm$0.36 & 95.15$\pm$0.14 & 87.62$\pm$0.59 & 88.23$\pm$2.12 & 93.83$\pm$1.39 & 87.71$\pm$2.04 \\
    MHIM-MIL~\cite{tang2023multiple} & 92.48$\pm$0.35 & 96.49$\pm$0.65 & 90.75$\pm$0.73 & 89.93$\pm$3.37 & 95.53$\pm$1.74 & 89.71$\pm$2.92 \\
    FiVE (Ours) & \textbf{94.25$\pm$0.33} & \textbf{97.56$\pm$0.59} & \textbf{93.24$\pm$0.72} & \textbf{94.62$\pm$2.13} & \textbf{96.33$\pm$1.21} & \textbf{93.89$\pm$1.90} \\
    \bottomrule
  \end{tabular}
  \caption{Comparison Performance of slide classification on Camelyon16 and TCGA Lung Cancer.}
  \label{tab_compare}
\end{table*}

Our model exhibites remarkable performance in zero-shot classification, achieving an accuracy of 71.26\%, even surpassing the SOTA method's one-shot experiment. Upon the introduction of training data, our models display exceptional transferability, outperforming the SOTA by 12.90\% in the one-shot setting.
The model's performance improves accordingly with the number of shot. Upon reaching 16-shot, our model reaches an impressive accuracy of 91.25\%, showcasing a notable 9.19\% improvement close to the fully supervised performance level.

\subsection{Performance Comparison with Existing Works}
We compared FiVE with ABMIL~\cite{ilse2018attention}, DSMIL~\cite{li2021dual}, CLAM-SB~\cite{lu2021data}, CLAM-MB~\cite{lu2021data}, TransMIL~\cite{shao2021transmil}, DTFD-MIL~\cite{zhang2022dtfd}, and MHIM-MIL~\cite{tang2023multiple}, all of which are attention-based MIL methods. In addition, we included two traditional MIL pooling operations, Max-pooling and Mean-pooling, for comparison. 
The results of all other methods are reproduced using the official code they provide under the same settings. 

To evaluate the performance of FiVE, we conducted fine-grained pre-training exclusively on the TCGA dataset, followed by fully supervised experiments on Camelyon16 and TCGA Lung Cancer datasets for WSI classification. Note that the test data is not used in pre-training. Results are shown in ~\cref{tab_compare}. It can be found that our method outperforms all the other baselines by a great margin, which fully demonstrates the significance of our fine-grained training scheme in improving performance on downstream tasks. Besides, since the pre-training data mainly comes from TCGA data, on the other hand, the task-specific prompt design is more suitable for TCGA data. This results in the performance improvement of our method for TCGA data being greater than Camelyon16 data.

\begin{table}[h]
	\centering
        \begin{tabular}{cccc|ccc}
          \hline
        SA & FGL & FGG & LDP & ACC & AUC & F1-score \\
          \hline
           \checkmark &  &  &  & 89.77 & 92.85 & 89.58 \\
           \checkmark & \checkmark &  &  & 91.21 & 94.36 & 91.01 \\
           \checkmark & \checkmark & \checkmark &  & 93.56 & 96.01 & 93.17\\
           \checkmark & \checkmark & \checkmark  & \checkmark & \textbf{94.62} & \textbf{96.33} & \textbf{93.89} \\
		\hline
	\end{tabular}
 	\caption{The ablation experiments of the pre-trained model fine-tuned on the TCGA Lung Cancer. SA, FGL, FGG, and LDP represent Self Attention, Fine-Grained Labels, Fine-Grained Guidance, and Learnable Diagnosis Prompts, respectively.}
	\label{tab_multimodal_diff_desc}
    \vspace{-1.0em}
\end{table}

\subsection{Ablation Studies}

\subsubsection{Effectiveness of TFS Module}
We focused on evaluating the impact of the TFS module on the model's overall performance with the TCGA Lung Cancer. The results detailed in ~\cref{tab_multimodal_diff_desc} reveal significant improvements at each stage of feature enhancement. Initially, the model with only Self Attention attained ACC of 89.77\%, AUC of 92.85\%, and F1-score of 89.95\%. The incorporation of fine-grained labels led to increases of 1.44\% in ACC, 1.51\% in AUC, and 1.43\% in F1-score. Subsequent integration of additional fine-grained guidance further improved performance. Ultimately, full framework with the TFS Module achieved the highest performance of 94.62\% ACC, 96.33\% AUC, and 93.89\% F1-score. These ablation experiments highlight the benefits of integrating these methods in the WSI classification task.

\subsubsection{Effectiveness of Patch Sample Strategy}
We verified the impact of different sampling strategies on model performance on TCGA data. Here, we assumed that there are two main indicators that affect model performance, sample ratio and max sample threshold (represented as MAXN). At the same time, in order to differentiate the experimental results, we used the unfrozen image encoder for experimental verification.


\begin{figure}[h]
  \centering
  \begin{subfigure}{0.493\linewidth}
    \includegraphics[width=\linewidth]{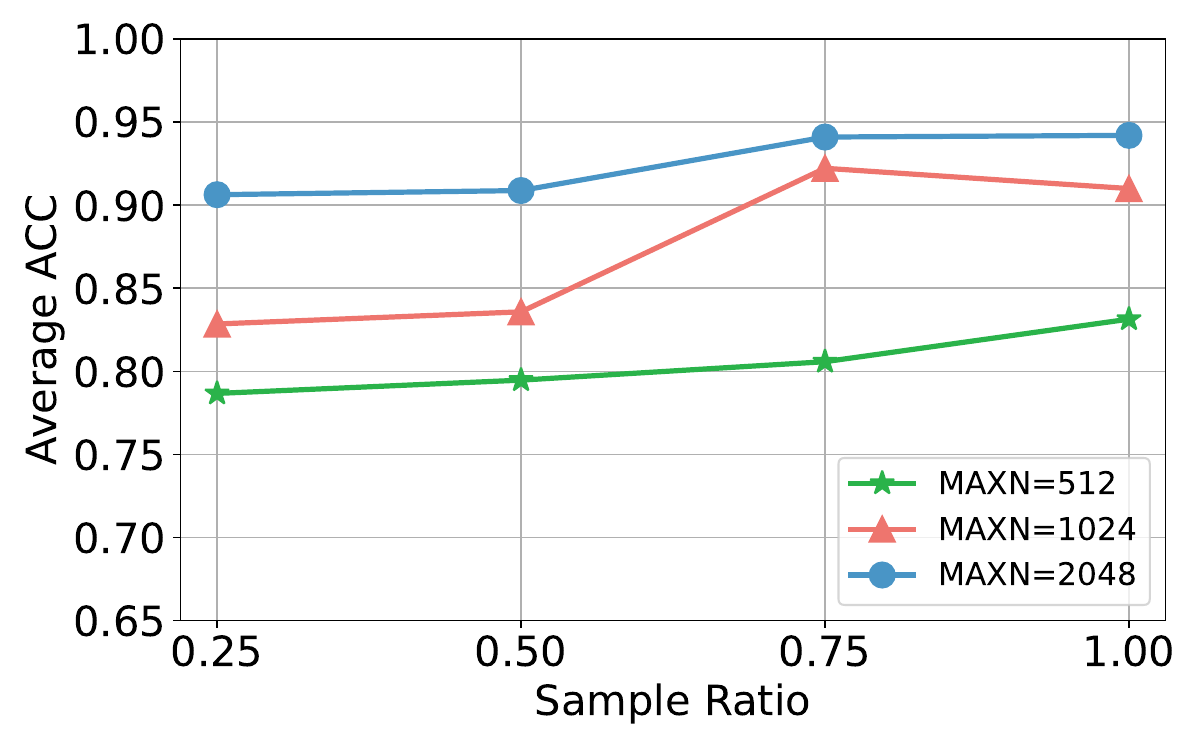}
    \label{fig:avg acc}
  \end{subfigure}
  \begin{subfigure}{0.493\linewidth}
    \includegraphics[width=\linewidth]{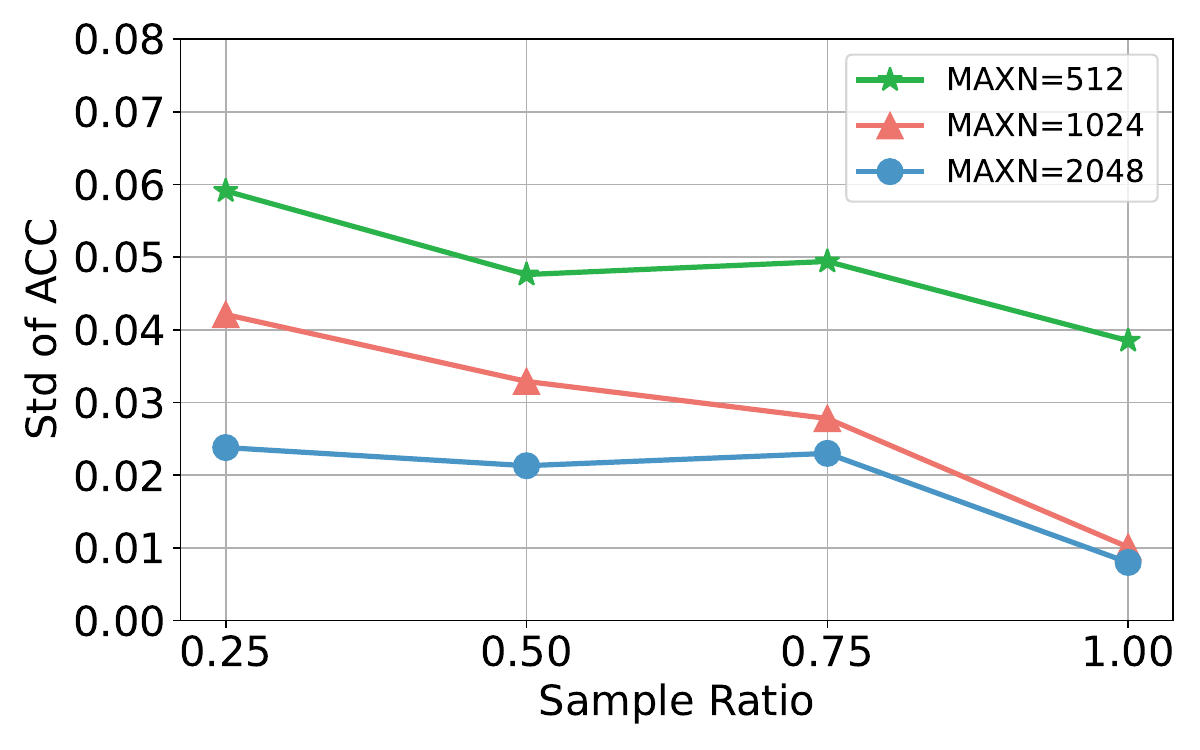}
    \label{fig:std acc}
  \end{subfigure}
  \vspace{-3.0em}
  \caption{Classification performance on TCGA Lung Cancer with diverse sampling strategies, presenting average and standard deviation (std) ACC values.}
  \label{fig_sample}
\end{figure}

As shown in ~\cref{fig_sample}, 
when MAXN is sufficiently large ($\geq 2048$), the correlation between the model's performance and the sample ratio is not significant. Even with a small sample ratio ($\leq 0.5$), the model can still effectively align with the original data distribution. Conversely, when MAXN is not large enough ($\textless 2048$), the primary change in the model's performance depends on whether the magnitude of the sample ratio can match the original data distribution. When the sample ratio increases to 0.75, a steep performance improvement can be observed, suggesting that the sampled data at this point conforms to the original data distribution, after which the model's performance stabilizes.



Based on comprehensive experimental results, we established the sample ratio and MAXN as 0.5 and 2048 as suitable hyperparameters for the unfrozen image encoder. As for the frozen image encoder experiment, taking into account the performance constraints imposed by the frozen image encoder on the model's performance upper limit~\cite{zhang2023text}, we recommend the sample ratio and MAXN to be set at 0.5 and 16384 based on our experiments.

\section{Conclusion}
In this paper, we introduce FiVE, a novel framework that demonstrates robust generalization and strong transferability for WSI classification. Our work pioneers the utilization of non-standardized WSI-report pairs from public databases to develop a VLM. 
To capture the complexities and diversity within these reports, we introduce the Task-specific Fine-grained Semantics (TFS) module. This module reconstructs fine-grained labels and diagnosis prompts during training, enhancing the semantic relevance of its features by introducing diagnosis prompts.
Furthermore, considering that pathological visual patterns are redundantly distributed across tissue slices, we employ a sampling strategy to reduce computational costs.
Our experiments demonstrate the robust generalizability and computational efficiency of the proposed framework, which can also be easily adapted to other tasks with minimal fine-tuning. We aspire to provide empirical insights and contribute to AI pathology research.

{
    \small
    \bibliographystyle{ieeenat_fullname}
    \bibliography{main}
}


\input{./supp}

\end{document}

%% file: preamble.tex
\usepackage[dvipsnames]{xcolor}


%% file: supp.tex
\clearpage
\setcounter{page}{1}
\maketitlesupplementary
\appendix

This supplementary material includes demonstrations of raw unstandardized pathology reports as presented in Sec.~\ref{report_demo}. 
And the detailed process of generating label descriptions using GPT-4, as well as specific label descriptions, are provided in Section~\ref{tcga_label}.
Additionally, further elaboration on the prompts utilized for Text Standardization is provided in Sec.~\ref{prompts}. 
Moreover, a comprehensive explanation of the Fine-Grained Guidance Construction is outlined in Sec.~\ref{fine_guidance}.

\section{Pathology Report Example}\label{report_demo}
\begin{figure*}[h]
  \centering
  \begin{subfigure}{0.33\linewidth}
    \includegraphics[width=\linewidth]{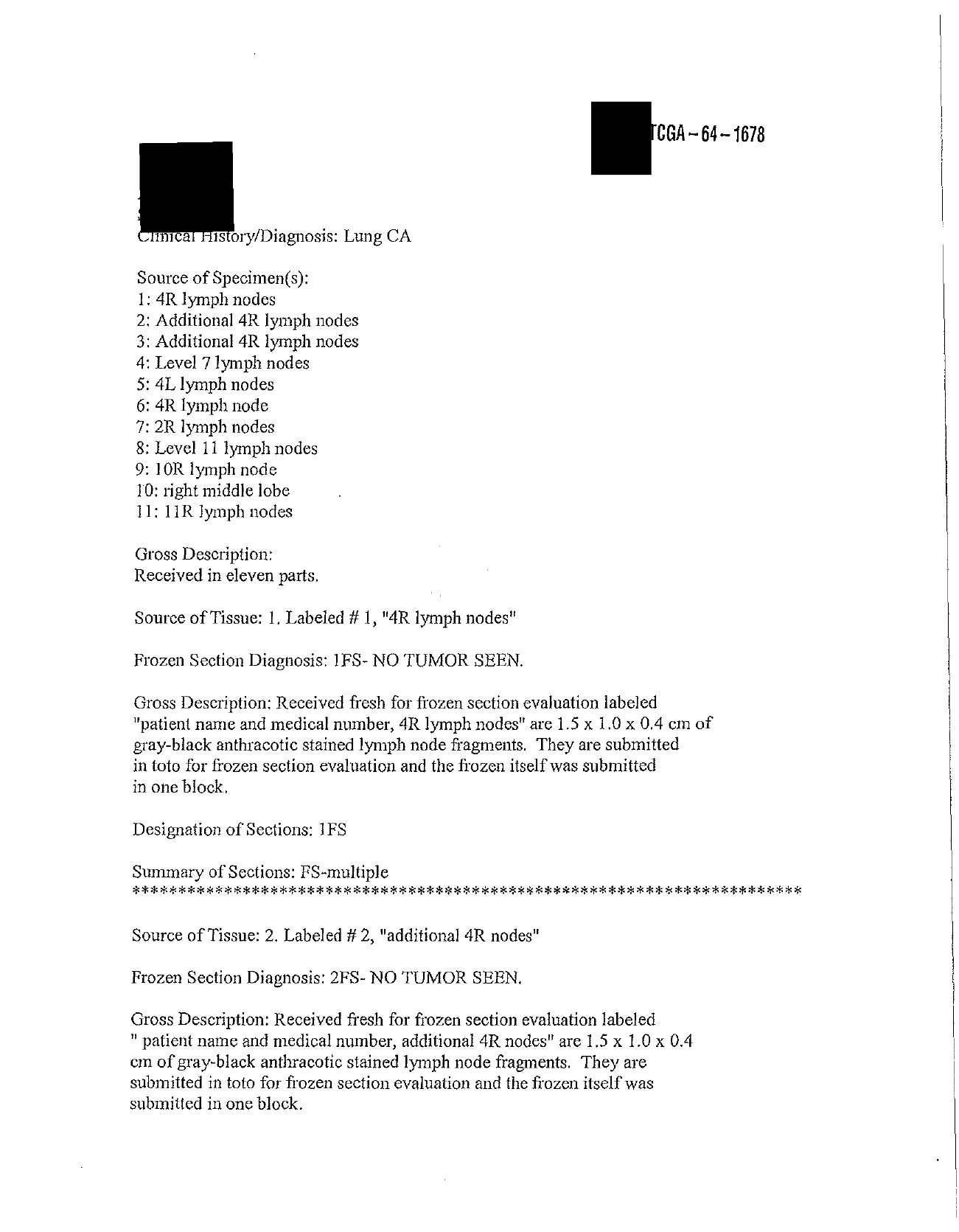}
    \caption{TCGA-64-1678}
    \label{fig:TCGA-64-1678}
  \end{subfigure}
  \begin{subfigure}{0.33\linewidth}
    \includegraphics[width=\linewidth]{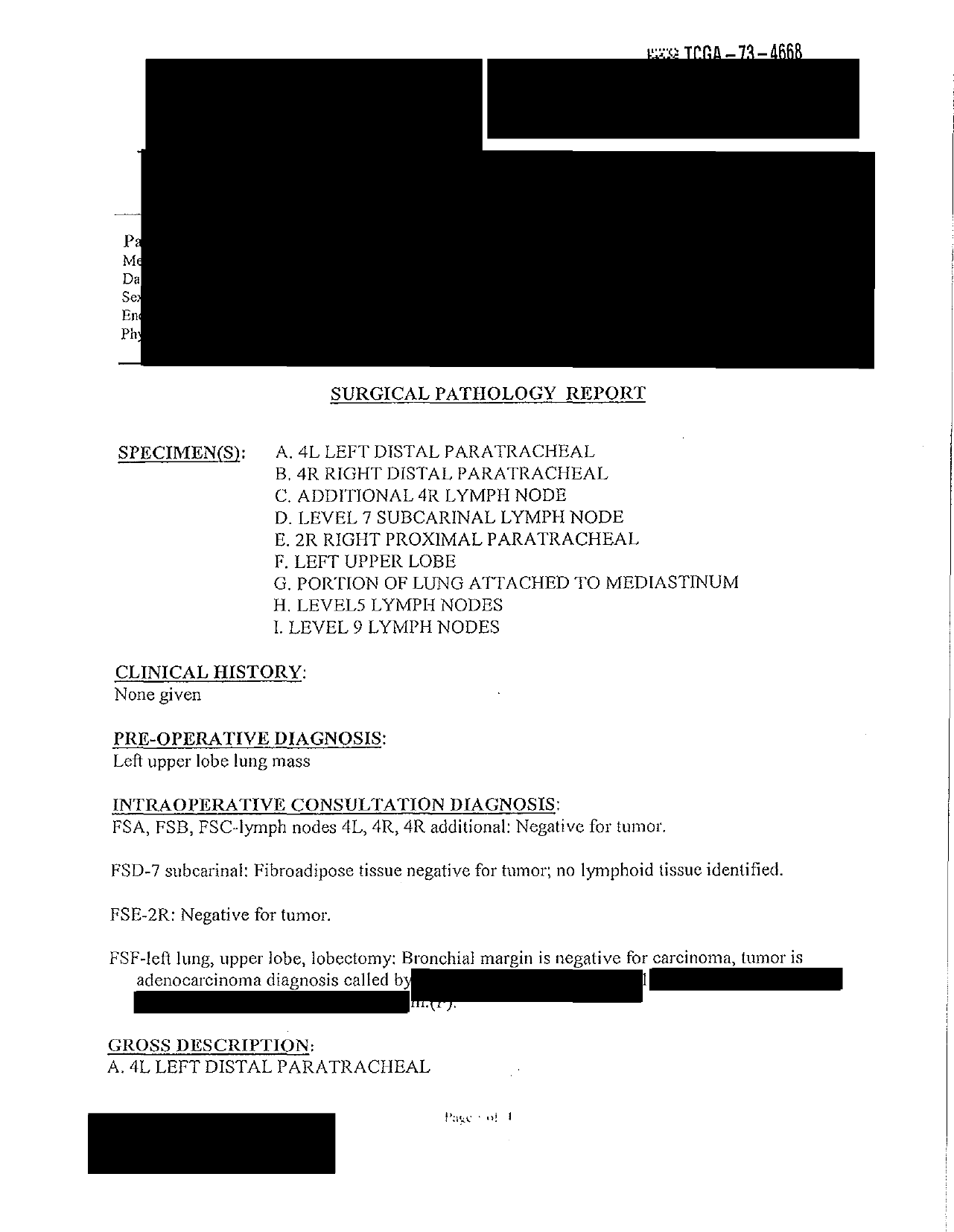}
    \caption{TCGA-73-4668}
    \label{fig:TCGA-73-4668}
  \end{subfigure}
    \begin{subfigure}{0.33\linewidth}
    \includegraphics[width=\linewidth]{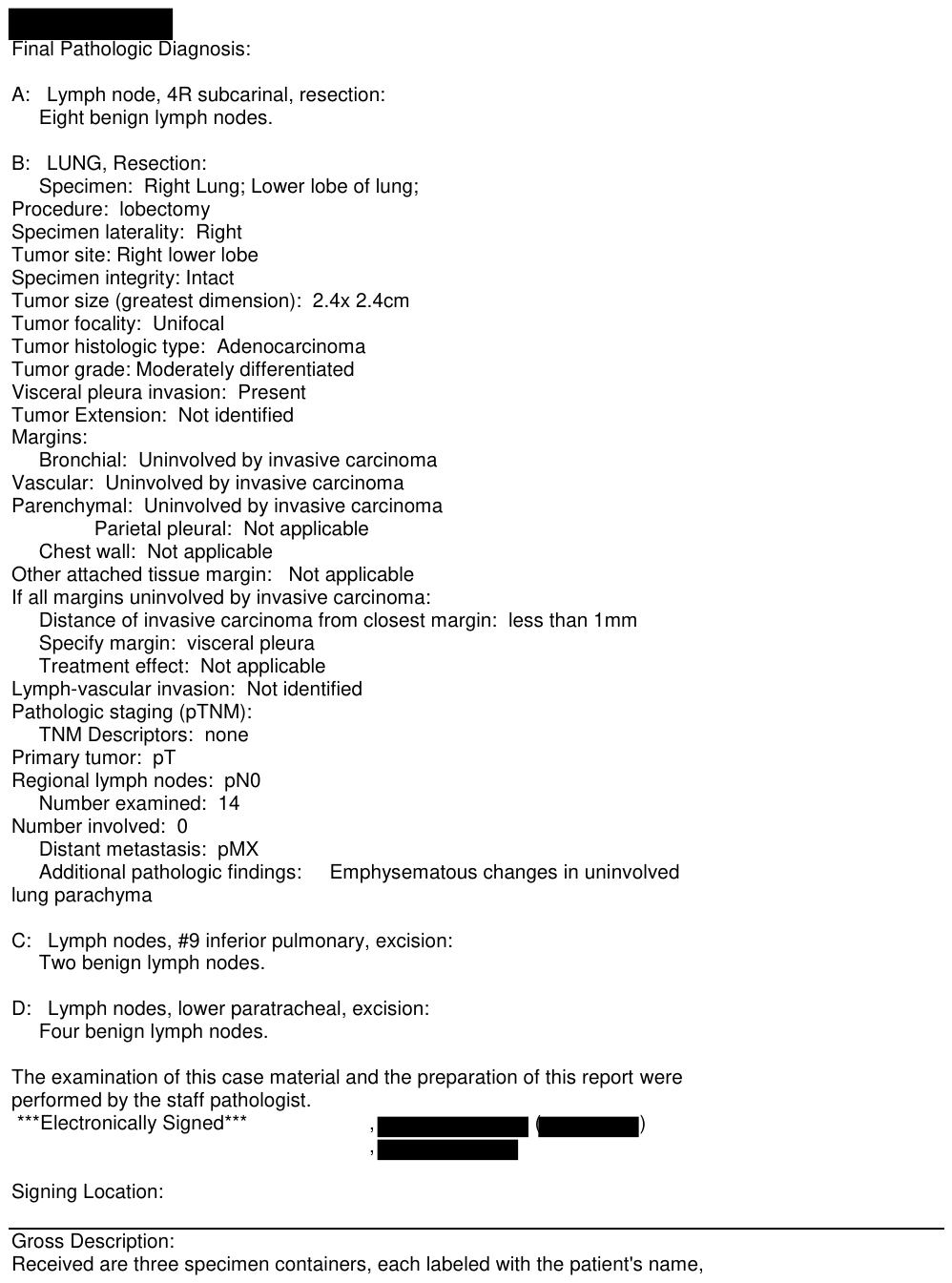}
    \caption{TCGA-91-6835}
    \label{fig:TCGA-91-6835}
  \end{subfigure}
  \caption{Pathology report examples. We randomly sample several pathology reports with different reporting standards for display. Sensitive information has been masked.}
  \label{fig_report_sample}
\end{figure*}

We present a range of pathology report demonstrations in ~\cref{fig_report_sample}, highlighting the extracted information utilizing GPT-4 to generate fine-grained text description label, as depicted in ~\cref{tab_fine_labels_example}. 
The sampled diagrams depict considerable variability in the formats of pathology reports stemming from diverse data sources. After the data extraction by GPT-4, the textual descriptions exhibit a more structured format. Although some inconsistencies in expression persist, our objective differs from multi-label classification, emphasizing the semantic expression within the text. This kind of comprehensive description at the slide level enhances our model's understanding of semantics, consequently refining the model's ability to generalize.

\begin{table*}[h]
	\centering
        \begin{tabular}{cc}
          \hline
        ID & Fine-Grained Text Description Label \\
          \hline
           TCGA-64-1678 & \makecell[l]{Differentiation of the lesion is poorly differentiated; Unknown; No indication of vascular invasion by\\ the lesion; No indication of pleural invasion by the lesion, as no visceral pleural invasion is seen;\\ Unknown; Margins of the excised tissue are clear of disease (R0).} \\\hline
           TCGA-73-4668 &  \makecell[l]{Lesion differentiation is moderately differentiated; Unknown; No indication of vascular invasion by\\ the lesion, as angiolymphatic invasion is absent; No indication of pleural invasion by the lesion, as\\ visceral pleural involvement is absent; Unknown; Margins of the excised tissue are clear of disease,\\ as the bronchial margins are uninvolved.}\\\hline
           TCGA-91-6835 &  \makecell[l]{Lesion differentiation is moderately differentiated; Unknown; No indication of vascular invasion by\\ the lesion, as vascular margins are uninvolved by invasive carcinoma; Pleural invasion by the lesion\\ is present; Unknown; Margins of the excised tissue are clear of disease, as all margins are uninvolved\\ by invasive carcinoma.}\\
		\hline
	\end{tabular}
	\caption{Fine-grained text description labels extracted from raw pathology reports.}
	\label{tab_fine_labels_example}
\end{table*}

\section{TCGA Label Description}\label{tcga_label}
We utilize the prompt ``Describe the morphological characteristics of the \textbf{LABEL} in a single sentence in English." to obtain label descriptions through GPT-4. 
When utilized, the placeholder tag \textbf{LABEL} is substituted with each specific label in the process.

\subsection{TCGA Lung Cancer Label}
\noindent \textbf{LUAD: }LUAD (Lung Adenocarcinoma) typically exhibits a diverse array of cell types, including glandular, papillary, and acinar structures with mucin production, and varying differentiation levels from well-differentiated to poorly-differentiated.

\noindent \textbf{LUSC: }LUSC (Lung Squamous Cell Carcinoma) is characterized by tumor cells forming sheet-like squamous structures, possibly showing keratinization features like keratin pearl formation, and is typically well-differentiated.

\subsection{TCGA-LUAD Subtype Label}
To evaluate the generalizability of our model and perform zero-shot classification, we curate a dataset of histological subtype labels for TCGA-LUAD , including histological subtypes of 54 samples of Lung Adenocarcinoma Mixed Subtype, 10 samples of Lung Bronchioloalveolar Carcinoma Nonmucinous, 5 samples of Lung Acinar Adenocarcinoma, 1 sample of Mucinous Adenocarcinoma, 4 samples of Micropapillary (colloid) Adenocarcinoma, 3 samples of Lung Bronchioloalveolar Carcinoma Mucinous, 2 samples of Lung Micropapillary Adenocarcinoma, 8 samples of Lung Papillary Adenocarcinoma. 


\noindent \textbf{Adenocarcinoma Mixed Subtype: } Adenocarcinoma mixed subtype is a cancer characterized by the presence of diverse cell types, exhibiting a combination of morphological features from various adenocarcinoma subtypes within the same tumor, making it a heterogeneous and challenging histological entity.

\noindent \textbf{Bronchioloalveolar Carcinoma Nonmucinous: } Bronchioloalveolar carcinoma nonmucinous is a type of lung cancer characterized by the proliferation of well-differentiated, nonmucinous glandular cells along the bronchiolar and alveolar structures within the lung tissue, often presenting as solitary nodules or lepidic growth patterns.

\noindent \textbf{Papillary Adenocarcinoma: } Papillary adenocarcinoma is a type of cancer characterized by finger-like projections or papillae composed of malignant glandular cells, often exhibiting a well-differentiated appearance under a microscope.

\noindent \textbf{Acinar Adenocarcinoma: } Acinar adenocarcinoma is a form of cancer marked by the presence of glandular structures resembling acini, often comprised of well-differentiated malignant cells that form small, round, or oval-shaped structures.

\noindent \textbf{Mucinous (colloid) Adenocarcinoma: } Mucinous (colloid) adenocarcinoma is a cancer subtype characterized by the presence of abundant extracellular mucin, produced by malignant glandular cells, giving it a gelatinous or colloid-like appearance when viewed under a microscope.

\noindent \textbf{Bronchioloalveolar Carcinoma Carcinoma: } Bronchioloalveolar carcinoma mucinous is a lung cancer subtype characterized by the proliferation of glandular cells producing abundant mucin, often leading to a lepidic growth pattern and presenting as a mass with a mucinous appearance.

\noindent \textbf{Mucinous Adenocarcinoma: }
Mucinous adenocarcinoma is a type of cancer characterized by the abundant production of mucin, a gel-like substance, by malignant glandular cells, often resulting in a tumor with a significant mucinous component.

\noindent \textbf{Micropapillary Adenocarcinoma: }
Micropapillary adenocarcinoma displays distinct small clusters or papillary structures with a central clear space, resembling small grape-like formations when observed under a microscope.

\section{Prompts Used for Text Standardization via GPT-4}\label{prompts}
We present the prompt template employed for GPT-4 to extract information from the raw pathology report, wherein the placeholder tag \textbf{REPORT} is substituted with each distinct raw pathology report. Additionally, besides providing essential prompt information, we have manually annotated some assistant examples to assist GPT-4 in further standardizing the output data format.

\subsection{Prompt Template}
Based on the diagnose report provided, 
please summarize the report briefly and academically from the following perspectives as a medical professional, 
answer in phrases or medical vocabulary entity whenever possible to save words, don't leave out important information. 
Connect the answers in one sequence, separated them with semicolons (important). 
Important notes: For all perspectives, focus on the microscopic description rather than gross description; Ignore the lymph nodes information; 
If can't answer from the specific perspective, just answer ``Unknown.'' without another words!!! 
1. What is the differentiation of the lesion? (maybe: Well-differentiated; Moderately differentiated; Poorly differentiated; Moderately to poorly differentiated; Mixed differentiation. or others.) 
2. Is there any indication of spread through air spaces around the lesion? and explain the reasons. 
3. Is there any indication of vascular invasion by the lesion? and explain the reasons. 
4. Is there any indication of pleural invasion by the lesion? and explain the reasons. 
5. Is there any evidence of the lesion invading adjacent tissues or organs (excluding the current lung organ)? and explain the reasons. 
6. Are the margins of the excised tissue clear of disease? (note that: R0 means negative; R1 R2 are both mean positive; Rx means Unknown, just answer `Unknown' only.).
Diagnose report: \textbf{REPORT}

\subsection{Assistant Example}
\noindent \textbf{Raw Report Example 1: }
Gross Description: Microscopic Description: Diagnosis Details: Comments: Formatted Path Reports: LUNG TISSUE CHECKLIST. Specimen type: Lobectomy. Tumor site: Lung. Tumor size: 6 x6x6cm. Histologic type: Squamous cell carcinoma. Histologic grade: Moderately differentiated. Tumor extent: Visceral pleura. Other tumor nodules: Not specified. Lymph nodes: 1/3 positive for metastasis (Intrathoracical 1/3). Lymphatic invasion: Not specified. Venous invasion: Not specified. Margins: Not specified. Evidence of neo-adjuvant treatment: Not specified. Additional pathologic findings: Not specified. Comments: Left-lower. to 11/8/12 l'es.

\noindent \textbf{Answer Example 1: }
Moderately differentiated; Unknown; Unknown; Pleural invasion indicated due to tumor extent to visceral pleura; Unknown; Unknown.

\noindent \textbf{Raw Report Example 2: }
REVISED REPORT (Revised information underlined). TISSUE DESCRIPTION: Tissue from the left kidney (partial nephrectomy, 3.76. grams, 2.2 X. 2.0 X 1.9 cm). . DIAGNOSIS: Kidney, left, partial nephrectomy: Grade 1 (of 4) renal. cell. carcinoma, papillary type, forms a 1.9 X 1.7 x 1.7 cm. mass. The tumor is confined to the kidney. Coagulative. tumor. necrosis is absent. Sarcomatoid differentiation is. absent. The. surgical margins are negative for tumor (free by 0.2 cm). AMENDMENTS. Revision Description: Review of permanent sections reveals the tumor to be a. grade 1 (of. 4) renal cell carcinoma, papillary type. Original Diagnosis. Kidney, left, partial nephrectomy: Grade 1 (of 4) renal. cell. carcinoma, clear cell type, forms a 1.9 X 1.7 X 1.7 cm. mass. The. tumor is confined to the kidney. Coagulative tumor. necrosis is. absent. Sarcomatoid differentiation is absent. The. surgical. margins are negative for tumor (free by 0.2 cm).

\noindent \textbf{Answer Example 2: }
Lesion is grade 1 (of 4), indicating well-differentiated; Unknown; Unknown; Unknown; The lesion is confined to the kidney, indicating no invasion of adjacent tissues or organs; Margins of the excised tissue are clear of disease (free by 0.2 cm).

\noindent \textbf{Raw Report Example 3: }
TISSUE DESCRIPTION: Right lower lobe lung (305 grams) superior (right upper and lower. paratracheal) and inferior (subcarinal, inferior pulmonary ligament). mediastinal lymph nodes and N1 (right interlobar) lymph nodes. DIAGNOSIS: Lung, right lower lobe, lobectomy : Grade 3 (of 4) squamous cell. carcinoma forming a subpleural mass measuring 5 X 4.5 X 2.6 cm,. extending into but not through the pleura. Bronchial margin is. negative for tumor. Multiple (6) intrapulmonary peribronchial lymph. nodes are negative for tumor. Lymph nodes, superior and inferior mediastinal, N1, excision: Multiple superior (2 right lower and 2 right upper paratracheal) and. inferior (6 subcarinal, 1 inferior pulmonary ligament) mediastinal. lymph nodes and N1 (3 right interlobar) lymph nodes are negative for. tumor.

\noindent \textbf{Answer Example 3: }
Lesion differentiation is poorly differentiated (G3 of 4); Unknown; Unknown; Pleural invasion by the lesion is present, as the carcinoma extends into but not through the pleura; Unknown; Margins of the excised tissue are clear of disease (R0).

\begin{figure}[h]
    \centering
    \includegraphics[width=\linewidth]{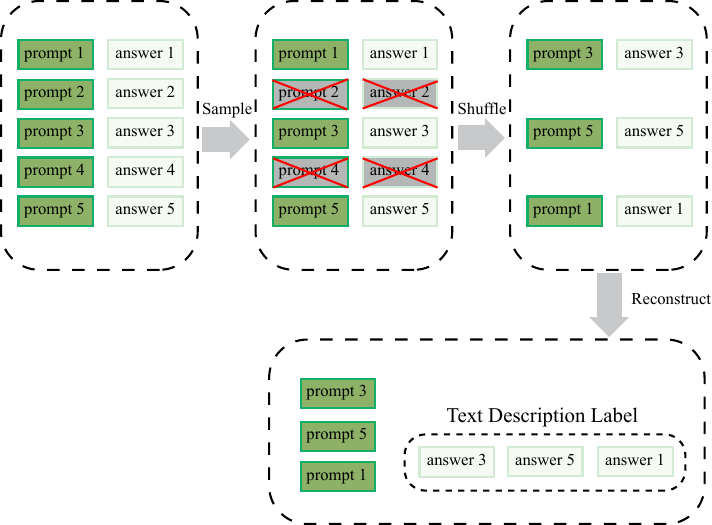} 
    \caption{ Fine-grained guidance construction pipeline.}
    \label{fig:Sample_Shuffle_Reconstruct}  \vspace{-1em}
\end{figure}

\section{Fine-grained Guidance Construction Pipeline}\label{fine_guidance}
As depicted in~\cref{fig:Sample_Shuffle_Reconstruct}, we partition the original fine-grained text descriptions into several parts based on manually designed prompts. Next, we perform random sampling and remove ``Unknown'' tags from the answers. Afterward, we shuffle the retained pairs and reconstruct the answers into a unified and complete sequence..
This approach enables our model to undergo a wider range of description transformations, with changes in granularity providing diverse perspectives on the visual image. Aligning visual images with text descriptions of varying granularities enriches the training data, offering more descriptive perspectives and semantic information for visual patterns. This enhances the semantic richness of the model's features and facilitates model transfer to downstream tasks.